\documentclass[sigconf, nonacm]{acmart}

\AtBeginDocument{%
  }

\setcopyright{none}
\copyrightyear{2026}
\acmYear{2026}
\acmDOI{XXXXXXX.XXXXXXX}

\usepackage{amsmath,amsfonts,bm}

\def\eqref#1{equation~\ref{#1}}

\def\1{\bm{1}}

\def\eps{{\epsilon}}

\def\vz{{\bm{z}}}

\DeclareMathAlphabet{\mathsfit}{\encodingdefault}{\sfdefault}{m}{sl}
\SetMathAlphabet{\mathsfit}{bold}{\encodingdefault}{\sfdefault}{bx}{n}

\newcommand{\R}{\mathbb{R}}

\usepackage{hyperref}
\usepackage{url}

\usepackage{dirtytalk}
\usepackage{xspace}
\usepackage{wrapfig}
\usepackage{algorithm}
\usepackage[noend]{algpseudocode}
\usepackage{graphicx}
\usepackage{lipsum}
\usepackage{circledsteps}
\usepackage{enumitem}

\newcommand\norm[1]{\left\lVert#1\right\rVert}

\renewcommand{\eps}{\varepsilon}

\newcommand{\gem}{$\operatorname{GEM+}$\xspace}
\newcommand{\bfgem}{$\operatorname{\textbf{GEM+}}$\xspace}

\newtheorem{theorem}{Theorem}[section]

\newtheorem{definition}[theorem]{Definition}

\begin{document}

\title{GEM+: Scalable State-of-the-Art Private Synthetic Data with Generator Networks}

\author{Samuel Maddock}
\authornote{Author correspondence to: smaddock@meta.com, shripadgade@meta.com}
\affiliation{%
  \institution{Meta Platforms, Inc.}
\city{London}
   \country{UK}
}
\author{Shripad Gade}
\authornotemark[1]
\affiliation{%
  \institution{Meta Platforms, Inc.}
  \city{Menlo Park}
  \state{CA}
  \country{USA}
}
\author{Graham Cormode}
\affiliation{%
  \institution{Meta Platforms, Inc.}
  \city{London}
   \country{UK}
}
\author{Will Bullock}
\affiliation{%
  \institution{Meta Platforms, Inc.}
\city{Menlo Park}
  \state{CA}
  \country{USA}
}

\renewcommand{\shortauthors}{Maddock et al.}

\begin{abstract}

State-of-the-art differentially private synthetic tabular data has been defined by adaptive “select–measure–generate” frameworks, exemplified by methods like AIM. These approaches iteratively measure low-order noisy marginals and fit graphical models to produce synthetic data, enabling systematic optimization of data quality under privacy constraints. Graphical models, however, are inefficient for high-dimensional data because they require substantial memory and must be retrained from scratch whenever the graph structure changes, leading to significant computational overhead. Recent methods, like GEM, overcome these limitations by using generator neural networks for improved scalability. However, empirical comparisons have mostly focused on small datasets, limiting real-world applicability. In this work, we introduce \gem, which integrates AIM’s adaptive measurement framework with GEM’s scalable generator network. Our experiments show that \gem outperforms AIM in both utility and scalability, delivering state-of-the-art results while efficiently handling datasets with over a hundred columns---where AIM fails due to memory and computational overheads.
\end{abstract}



\maketitle

\section{Introduction}
\label{sec:intro}

Differentially private synthetic data generation (DP-SDG) has become an essential approach for sharing sensitive tabular data while preserving individual privacy. By generating artificial data that retains the statistical properties of the source data and providing formal privacy guarantees, DP-SDG enables organizations to share, analyze, and train models on data without exposing sensitive information \citep{hardt2012simple, cormode2025synthetic}. State-of-the-art approaches achieve Differential Privacy (DP) by taking low-dimensional measurements, such as marginal queries, and adding calibrated noise. These noisy statistics are then used to train generative models, including graphical models \citep{zhang2017privbayes, mckenna2019graphical} and neural network-based generators \citep{liu2021iterative}. However, as most web-scale real-world datasets have a large number of rows and columns, the scalability of DP-SDG methods has become a critical concern. In online advertising, machine learning models often use hundreds of features, highlighting the need for scalable DP-SDG solutions to build private web advertising solutions \citep{sebbar2025criteoprivateads}.

The widely used \say{select–measure–generate} framework for private synthetic tabular data iteratively selects the worst-approximated query from a predefined workload (marginal queries), measures it with calibrated noise, and updates the generative model. This cycle incrementally improves utility for a target workload while ensuring differential privacy guarantees \cite{hardt2012simple}. AIM, the current state-of-the-art, presents an adaptive selection strategy for noisy marginal measurement followed by training a probabilistic graphical model \citep{mckenna2022aim, mckenna2019graphical}. AIM's adaptive selection is both workload and privacy aware, optimizing utility given privacy budget constraints. However, graphical models struggle with scalability: high-dimensional marginals change the structure of the graph by creating large cliques, increasing memory overheads and forcing retraining at each step. Additional heuristics are used to reduce memory by restricting the set of marginals that can be selected, which can harm utility. As a result, AIM is impractical for high-dimensional, real-world datasets.

To overcome these scalability issues, Liu et al. introduced GEM, which replaces graphical models with generator networks that can be incrementally updated without retraining from scratch~\citep{liu2021iterative}. However, GEM lacks AIM’s adaptive selection and privacy budget allocation strategies, resulting in lower utility. Moreover, most studies, including GEM, have only evaluated performance on low-dimensional datasets. Recent benchmarks \citep{chen2025benchmarking, ganev2023understanding} focus on data with up to 40 columns and 150,000 rows, leaving real-world scalability largely unexplored. As noted by Ganev et al.~\citep{ganev2024graphical}, while deep learning approaches may offer advantages at scale, empirical evidence is still lacking.

\noindent
\textbf{Our Contributions.} %
We present \gem, which integrates AIM's adaptive measurement framework with GEM's scalable generator neural networks, along with algorithmic improvements tailored for high-dimensional data. Our contributions are twofold:

\textbf{1.} We demonstrate that \gem breaks through the scalability barriers of AIM, efficiently handling datasets with hundreds of columns where AIM fails due to memory and computational constraints. On the Criteo Ads dataset \cite{sebbar2025criteoprivateads}, featuring up to 120 columns and 2.5 million rows, AIM cannot operate beyond 60 columns, whereas \gem seamlessly scales to the full 120 columns.

\textbf{2.} We show the superior utility of \gem, against the current state-of-the-art AIM, on tables where both AIM and \gem run tractably, setting a new benchmark for differentially private synthetic data generation.

\section{Select-Measure-Generate Framework}
\label{sec:smg}

\textbf{Differential Privacy.} %
We focus on synthetic data generation methods that ensure
$(\eps, \delta)$-Differential Privacy for sensitive datasets \cite{dwork2006calibrating}, a strong and widely accepted privacy definition. In this work, we adopt an alternative but equivalent formulation: zero-Concentrated Differential Privacy (zCDP)~\citep{bun2016concentrated}.
\begin{definition}[$\rho$-zCDP] \label{def:zcdp}
An algorithm $\mathcal{M}$ is $\rho$-zCDP if for any two neighboring datasets $D, D^\prime$ and all $\alpha \in (1,\infty)$ we have $D_\alpha(\mathcal{M}(D) | \mathcal{M}(D^\prime) \leq \rho \cdot \alpha$,
where $D_\alpha$ is R\'enyi divergence of order $\alpha$.
\end{definition}
\noindent zCDP has two key properties: composition and post-processing. Composition implies running two zCDP algorithms with parameters $\rho_1$ and $\rho_2$ results in $(\rho_1 + \rho_2)$-zCDP guarantees. Post-processing ensures that any function applied to the output of a $\rho$-DP algorithm preserves the same privacy guarantee. In practice, final privacy guarantees are reported as $(\eps, \delta)$-DP, which can be converted to and from $\rho$-zCDP guarantees \citep{bun2016concentrated}. \\

\noindent \textbf{Select-Measure-Generate.} The \say{select-measure-generate} framework defines a workload of queries $W$ with the goal of generating a synthetic dataset $\tilde D$ that is $(\eps,\delta)$-DP and provides accurate answers for all $q \in W$. The high-level procedure is outlined in Algorithm~\ref{alg:main}. This framework relies on two key DP mechanisms.

\textbf{Select: } The ExpMech is used to implement the ``select'' step. By defining $s(q) := \norm{q(D) - q(\tilde D)}$, the mechanism (approximately) selects the query $q^* \in W$ that is currently worst-approximated by the synthetic data model.
\begin{definition}[Exponential Mechanism (ExpMech)]
Let $s(q) : \mathcal{D} \rightarrow \R$ be a score function defined over a set of candidates $W$. The exponential mechanism releases $q$ with probability $\mathbb{P}[\mathcal{M}(D) = q] \propto \exp(\frac{\tau}{2\Delta} \cdot s(q))$,
with $\Delta := \max_q \Delta_1(s(q))$. This satisfies $\frac{\tau^2}{8}$-zCDP.
\end{definition}
\textbf{Measure:} Once a query is selected, it must be measured privately on the real data. This is achieved using the Gaussian mechanism:
\begin{definition}[Gaussian Mechanism]
Let $q: \mathcal{D} \rightarrow \R^d$ be a sensitivity $1$ query.
The Gaussian mechanism releases $q(D) + \mathcal{N}(0, \sigma^2 I_d)$ and satisfies $\frac{1}{2\sigma^2}$-zCDP.
\end{definition}

By combining ExpMech for selection and the Gaussian mechanism for measurement, each ``select-measure'' step incurs a total privacy cost of $\rho := \frac{1}{2\sigma^2} + \frac{\tau^2}{8}$ under zCDP. Since zCDP composes linearly, the total privacy cost for $T$ rounds of the ``select-measure-generate'' framework is simply $T\rho$. Given a total privacy budget $(\eps, \delta)$, one can work backwards to calibrate the noise parameters $\sigma, \tau$ by converting $(\eps,\delta)$ to a total $\rho$-zCDP budget and calibrate the per-round noise over $T$ rounds.

\textbf{Model Update \& Generate:} The measurements are used to train a generative model, $\theta$, commonly a graphical model or a generator network. As the model update uses noisy marginals only, it is a form of post-processing and incurs no additional privacy cost.
\begin{algorithm}[t]
    \caption{Select-Measure-Generate Framework \cite{hardt2012simple}\label{alg:main}}
    \begin{algorithmic}[1]
        \Require Private dataset $D$, workload of queries $W$, training rounds $T$, privacy parameters $(\eps, \delta)$
        \State Preprocess the workload $W^* := f(W)$
        \State Initialize %
        noise parameters $\sigma_t, \tau_t := g(\eps, \delta)$ via allocation $\alpha$
        \For{$t=1, \dots, T$}
            \State \textbf{Select:} via ExpMech $q_t \in W^*$ using score $s(q)$ and $\tau_t$
            \State \textbf{Measure:} marginal $q_{t}$ via $\tilde q_t = q_t(D) + N(0, \sigma_t^2)$
            \State \textbf{Update:} synthetic model $\theta_{t+1} := h(\theta_t, \{\tilde q_1, \dots, \tilde q_{t}\})$
            \State \textbf{Generate:} $\tilde D_{t} \sim \theta_{t}$
        \EndFor
        \State \textbf{Output:} $\tilde D \sim o(\{\theta_t\}_{t=1}^T)$
    \end{algorithmic}
\end{algorithm}
\subsection{Existing Methods: GEM and AIM}
\label{sec:related_work}

AIM and GEM follow the Select-Measure-Generate framework outlined in Algorithm \ref{alg:main}.
We summarize the individual decisions taken for the preprocessing function $f$, the per-round noise initialization function $g$, the selection function $s(q)$, the model training function $h$ and the post-processing function $o$. \\

\noindent \textbf{GEM \cite{liu2021iterative}:} GEM broadly follows the canonical approach of MWEM~\cite{hardt2012simple}, with the key idea of fitting a generator neural network based on the currently observed noisy measurements.
More specifically, at the start of training the workload of queries $W$ is left unchanged, $f(W) := W$,  fixed to that of $3$-way marginals. The budget initialization function $g(\eps, \delta)$ sets $\alpha = 0.5$ which spends an even amount of the total zCDP $\rho$ on selection and measurement and is calibrated to $T$ rounds of training.
The selection score $s(q)$ is simply the $L_1$ error i.e., $s(q) := \norm{q(D) - q(\tilde D_{t-1})}$ between source and synthetic data. The model update function, $h(\cdot)$, trains a generator neural network via a number of SGD steps on the current model $\theta_{t-1}$ to produce $\theta_{t}$.
The gradients for these updates are computed from the average $L_1$ loss between the current marginals produced by model $\theta_t$ and the observed noisy measurements $\{\tilde q_1, \dots, q_t\}$.
Finally, the post-processing function $o$ applies an Exponential Moving Average (EMA) over the last $T/2$ generator networks to get the final model. \\

\noindent \textbf{AIM \cite{mckenna2022aim}: } AIM initializes a workload, $W^* := f(W)$, which contains the downward closure of $W$ (i.e., all lower order marginals that can be formed from queries in $W$ are also added to $W^*$). The noise initialization function $g$ assigns $\alpha=0.9$ i.e., 90\% of the privacy budget to measurement ($\sigma_t$). The number of rounds are initialized to $T=16d$ where $d$ is the total number of columns. During training, an annealing condition is checked based on how much the last marginal measurement has changed between models. If the model estimate did not change by much, the noise parameter $\sigma$ is reduced by half.
The selection score used is $s_{\text{AIM}}(q) :=  \|q(D) - q(\tilde D)\| - \sqrt{2/\pi}\sigma n_{q_t}$ which adapts the GEM score to include an expected error term based on measuring $q$ under Gaussian noise. The model function $h(\cdot)$ trains a graphical model via the Private-PGM algorithm \cite{mckenna2019graphical}, this includes heuristics which prevent selecting marginals that may cause the graphical model to explode in memory due to a large clique forming. Last, the function $o(\cdot)$ returns the final graphical model trained on all measurements.

\section{Our Algorithm: \bfgem}
\label{sec:gem+}

Our approach, \gem (Algorithm \ref{alg:gemplus}), introduces several key modifications to the GEM algorithm, incorporating components from AIM along with additional enhancements for high-dimensional data.

\textbf{1. Workload Closure.}
GEM is defined to operate on a fixed workload of $3$-way marginals. Using the AIM framework, \gem extends this to support any provided target workload. As in AIM, we also form the closure of the workload: for any $k$-way marginal in $W$, all lower-order marginals that can be formed from it are included in the workload for (possible) selection during training.

\textbf{2. Initialization and Budget Allocation.} In the GEM algorithm the privacy budget is split evenly between selection and measurement. In \gem we follow the allocation and initialization strategy of AIM. This initializes $T=16d$ and sets $\alpha=0.9$ to spend 90\% of a round's budget on measurement with remainder (10\%) used for marginal selection. As in AIM, \gem measures all $1$-way marginals (under DP) to initialize the generator network before training begins, which does not occur in original GEM.

\textbf{3. Select Step.} In the original GEM algorithm, the select and measurement steps follow the MWEM approach \citep{hardt2012simple}, using a standard exponential mechanism with selection function $s(q) : = \norm{q(D) - q(\tilde D)}$. In \gem, we replace this with the more sophisticated selection strategy used by AIM, which enables more effective allocation of the privacy budget.

\textbf{4. Measurement Step.} The GAN in \gem is trained on all marginal vectors measured so far. This is unlike the original GEM, which uses pointwise marginals, \gem exclusively uses marginal vectors and eliminates the use of pointwise thresholding and top-$k$ updates. %

\begin{algorithm}[t]
\caption{GEM+}
\label{alg:gemplus}
\begin{algorithmic}[1]
\Require Private dataset $D$, workload $W$, privacy parameters $(\eps, \delta)$
\State %
\textbf{Workload Closure:} $W^* \gets$ downward closure of $W$
\State %
\textbf{Initialize:} Set $\alpha=0.9$ and initialize noise parameters $\sigma_t, \tau_t$ with $T=16d$. Measure 1-way marginals in $W^*$ via $\sigma_t$ to initialize $\theta_0$
\For{$t = 1$ to $T$}
    \State  %
    \textbf{Select:} Use ExpMech with $s_{\text{AIM}}(q)$ to select $q_t \in W^*$
    \State %
    \textbf{Measure:} $\tilde q_t = q_t(D) + N(0, \sigma_t^2)$
    \State %
    \textbf{Filter:} Remove $q_t$ from $W^*$ %
    \State %
    \textbf{Marginal Closure:} For each $\tilde q' \subseteq \tilde q_t$, estimate $\tilde{q}'_t$ for no extra privacy cost; add all new $\tilde{q}'_t$ to $\mathcal{M}$
    \State \textbf{Update:} Update generator $\theta_t$ by minimizing
        \[
        \mathcal{L}(\theta_t) = \frac{1}{|\mathcal{M}|} \sum_{ \tilde{q} \in \mathcal{M}} \left\| q(\theta_t(\vz)) - \tilde{q} \right\|_1
        \]
        where $\vz \in \R^{B\times d}$ is a fixed batch of gaussian noise, $q(\theta_t(\vz))$ denotes the soft marginal computed on the generator's output. %
    \State \textbf{Anneal:} If $||q_t(\theta_t(\vz)) - q_{t}(\theta_{t-1}(\vz))|| \leq \gamma$ then anneal $\sigma_t,\tau_t$
\EndFor
\State \textbf{Output:} $\tilde{D} \sim \theta_T$
\end{algorithmic}
\end{algorithm}

\textbf{5. Candidate Filtering.}
In original GEM, there is no condition that prevents repeated measurements from happening.
We found that GEM will often select many of the same marginals throughout training, which wastes privacy budget in high-dimensional scenarios and limits exploration. To mitigate this, in \gem, we introduce a filtering condition which prevents repeated selection of the same marginal if it has been previously measured. %

\textbf{6. Disabling EMA Averaging.} We disable the Exponential Moving Average (EMA) of the last $T/2$ generator networks that was proposed in GEM and see consistent performance improvements.

\textbf{7. Marginal Closure.} In \gem, when measuring a 3-way marginal with noise we also estimate all lower order marginals that are contained within that marginal for no additional privacy cost. With a marginal, $(A,B,C)$, via marginalization we can obtain estimates for $(A,B), (A,C), (B,C), (A), (B), (C)$. If the same measurement occurs multiple times, we take a weighted average using $1/\sigma$ as the weight. This is similar to how the Private-PGM algorithm handles repeated measurements in AIM. We observed this gives a small improvement to utility for \gem.

\textbf{8. + 9. Generator Training Improvements.}
GEM/\gem maintain a batch $B$ of latent random noise which is used as input to the generator network to sample synthetic data.
We found increasing both the number of SGD iterations ($t$) used to train the generator per round and increasing the latent batch size $B$ is crucial for good performance on large data.
We implement an early stopping threshold to end the current round of SGD training if the loss does not change significantly, preventing overfitting to noisy marginals.

\begin{table}[b]
    \caption{\gem ablation on Criteo with $d=30, \eps=1$}
    \vspace{-0.3cm}
    \label{tab:ablation}
    \begin{tabular}{l |l }
         \textbf{Method} ($T=$ rounds, $t = $ SGD iters, $B = $ batch size) & \textbf{$L_1$ error} \\
         \hline
         AIM (Benchmark)  ($T=16d, t=1000$) & 0.0904 \\
         \hline
         GEM (Benchmark) ($T=100, t=100, B=1000$) &  0.437\\
         GEM (Tuned) ($T=200, t=1500, B=25000$)&  0.261 \\
         \hline
         \gem = \Circled{1} + \Circled{2} + \Circled{3} + \Circled{4} + \Circled{5} %
         & 0.254 \\
         \hspace{3mm} + \Circled{6} w/o EMA & 0.101 \\
         \hspace{3mm} + \Circled{7} Marginal closure & 0.0976 \\
         \hspace{3mm} + \Circled{8} Increased latent batch $(B=25000)$ & 0.0579 \\
         \hspace{3mm} + \Circled{9} Increased SGD steps ($t=1500$) & \textbf{0.0573}
    \end{tabular}
\vspace{-0.3cm}
\end{table}

\section{Experimental Results}
\label{sec:exp}

\textbf{Setup.} We conduct our experiments using a recent online advertising dataset from Criteo \cite{sebbar2025criteoprivateads}. The dataset is partitioned by day with 150 columns in total. For our analysis, we select an entire day's data with approximately 2.5 million rows.
After removing columns that are IDs or arrays, we retain 123 columns. Missing (null) values are zero-imputed and numerical columns are uniformly discretized into $32$ bins, see Section \ref{sec:discussion} for more discussion. We compare our proposed \gem method with AIM and GEM. %
We use the original implementation by the authors of AIM and GEM and take hyperparameter guidance from the recent study of Chen et al. \cite{chen2025benchmarking}. To assess utility, we report the average $L_1$ workload error on  synthetic data generated from 3 independent training runs. We train AIM on CPU and GEM/\gem on a single A100 40GB GPU. \\

\begin{figure}[t]
    \centering
    \includegraphics[width=0.35\textwidth]{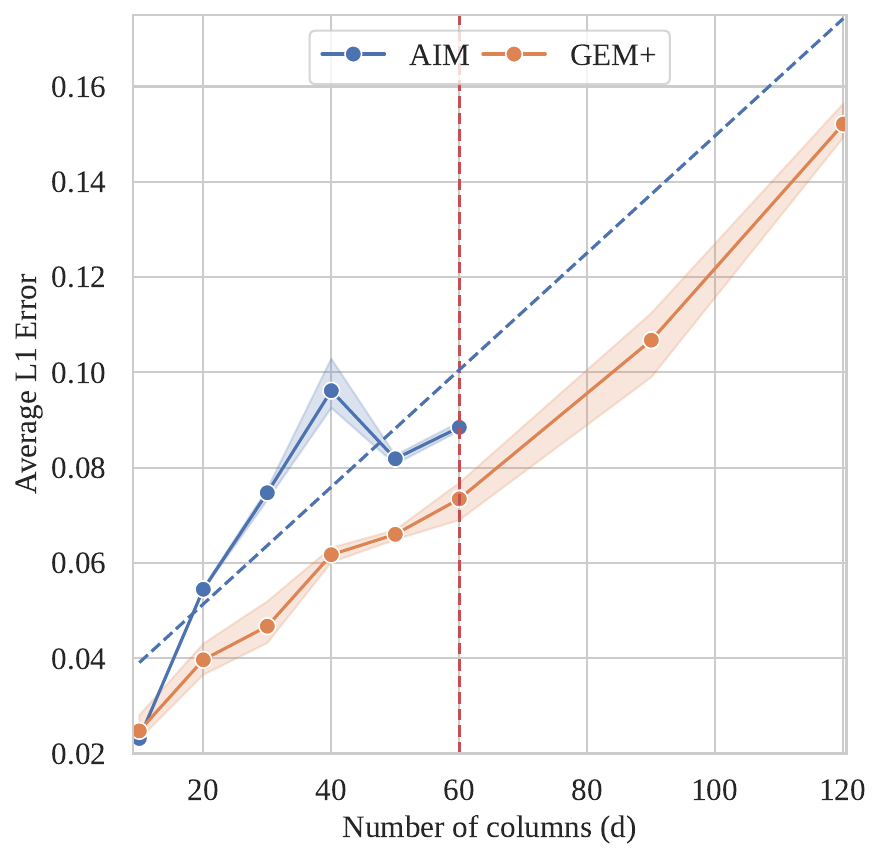}
    \vspace{-0.3cm}
    \caption{\gem vs. AIM on Criteo with $\eps=3$. We vary the number of columns $d \in \{10, 20, \cdots, 120\}$. \label{fig:1}} %
\end{figure}

\noindent\textbf{Comparing GEM with \bfgem.} We compare the original GEM with \gem in Table \ref{tab:ablation}, using $d=30$ and $\eps = 1, \delta = 10^{-6}$ on Criteo dataset. The original GEM, tested with both default and tuned hyperparameters, shows poor utility, with only slight improvement after tuning. In contrast, untuned \gem — combining AIM's adaptivity and GEM's generator, achieves lower error than tuned GEM. With further enhancements (Section \ref{sec:gem+}), the $L_1$ error drops to ~4x lower than GEM. Notably, a large latent noise batch size ($B$) is crucial for utility, contradicting GEM’s authors who fixed $B=1000$ as they did not see gains. Due to poor utility, we exclude the original GEM from further experiments. \\

\noindent\textbf{Scaling Synthetic Data Generation.} In Figure \ref{fig:1}, we vary the number of columns on Criteo from $10$ up to $120$ for AIM and \gem with $\eps = 3$. We impose a strict computational limit of $7$ days for any method. When $d \leq 60$, we observe \gem outperforms AIM in overall $L_1$ workload error. For $d > 60$, we find AIM encounters computation or out-of-memory errors and cannot complete training in this regime, whereas \gem successfully scales to 120 columns. \\

\begin{figure}[b] %
    \centering
    \includegraphics[width=0.3\textwidth]{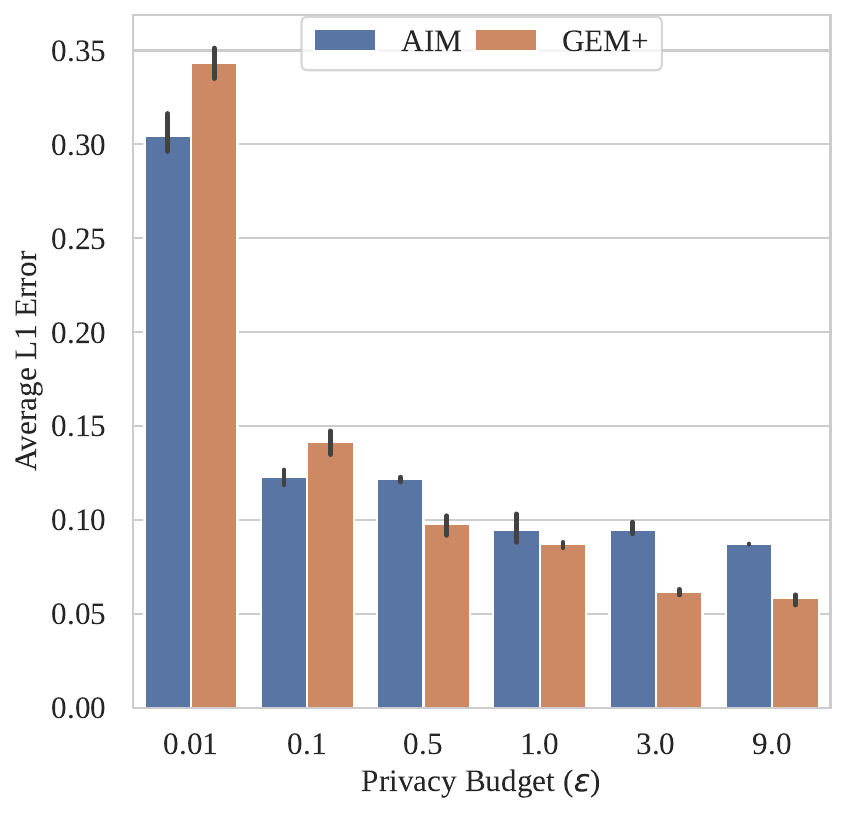}
    \vspace{-0.3cm}
    \caption{Varying the privacy budget ($\eps$) with $d=40$ \label{fig:2}}
    \vspace{-0.3cm}
\end{figure}

\noindent\textbf{Varying the privacy budget ($\eps$).} In Figure \ref{fig:2}, we vary the privacy budget $\eps$ and compare AIM and \gem with $d=40$. Two regimes emerge: for $\eps \geq 0.5$, \gem outperforms AIM, while for $\eps \leq 0.1$, AIM outperforms GEM. This indicates the neural-network backend of \gem is particularly effective in lower-noise scenarios, where it can better capture the information in the measured marginals. At reasonable privacy regimes ($\eps = 3$), we consistently demonstrate better quality (lower $L_1$ error) than AIM, demonstrating SOTA performance (Figure \ref{fig:1}). \\

\noindent\textbf{Runtime Comparisons.} In Table \ref{tab:runtime}, we report the average runtime of AIM and \gem with $\eps=3$ and $T=16d$. We clearly observe the exponential increase to AIM's computation time as the number of columns increases, exceeding 5 days of compute when $d=60$. In contrast, \gem scales steadily with the number of columns enabling efficient training even as the number of columns grows.

\section{Discussion}
\label{sec:discussion}

\noindent\textbf{Modeling flexibility of \bfgem:} A key advantage of \gem over AIM is the flexibility of increasing model capacity. %
The generator network backbone allows practitioners to easily adjust hyperparameters such as the training algorithm, latent batch size and model size to balance utility and training speed accordingly. This is very valuable in practical large dataset scenarios. Additionally, \gem naturally leverages GPU acceleration and can benefit from pretraining on public marginals to further improve performance. This is in contrast to AIM's graphical model which is less adaptable. \\

\noindent\textbf{Discretization in practice.} In our experiments, we discretized numerical columns into 32 bins to ensure AIM's tractability for 60 columns. However, the scalability of \gem (being a fixed model size) means it can accommodate even larger histogram sizes without prohibitive memory issues unlike AIM. This allows for finer-grained discretization to improve the fidelity of synthetic data, especially in applications where preserving numerical distributions is important. \\

\begin{table}[t]
    \caption{Training time for AIM $\&$ GEM. Criteo, $\eps=3, T=16d$.}
    \vspace{-0.3cm}
    \label{tab:runtime}
    \centering
    \begin{tabular}{c |c| c| c | c | c}
         \textbf{Method / Columns} & \textbf{20} & \textbf{40} & \textbf{60} &
         \textbf{90} &
         \textbf{120}
         \\
         \hline
         AIM & 11hrs & 25hrs & 129hrs & OOM &  OOM \\
         \gem & 5hrs & 25hrs & 32hrs & 39hrs & 41hrs
    \end{tabular}
    \vspace{-0.3cm}
\end{table}

\noindent\textbf{Conclusion.} Our results highlight SOTA performance of \gem. It also demonstrates the importance of evaluating DP-SDG methods on large, high-dimensional datasets that reflect real-world use cases, and we encourage the community to adopt datasets like Criteo for benchmarking.

\bibliographystyle{ACM-Reference-Format}
\bibliography{sample-base}

\end{document}